# Towards a Safer and Sustainable Manufacturing Process: Material classification in Laser Cutting Using Deep Learning

Mohamed Abdallah Salem, Hamdy Ahmed Ashur, Ahmed Elshinnawy

Smart Control Systems for Energy Management,
Arab academy for science, technology and maritime transport, Alexandria, Egypt

*Abstract*—Laser cutting is a widely adopted technology in material processing across various industries., but it generates a significant amount of dust, smoke, and aerosols during operation, posing a risk to both the environment and workers' health. Speckle sensing has emerged as a promising method to monitor the cutting process and identify material types in real-time. This paper proposes a material classification technique using a speckle pattern of the material's surface based on deep learning to monitor and control the laser cutting process. The proposed method involves training a convolutional neural network (CNN) on a dataset of laser speckle patterns to recognize distinct material types for safe and efficient cutting. Previous methods for material classification using speckle sensing may face issues when the color of the laser used to produce the speckle pattern is changed. Experiments conducted in this study demonstrate that the proposed method achieves high accuracy in material classification, even when the laser color is changed. The model achieved an accuracy of 98.30% on the training set and 96.88% on the validation set. Furthermore, the model was evaluated on a set of 3000 new images for 30 different materials, achieving an F1-score of 0.9643. The proposed method provides a robust and accurate solution for material-aware laser cutting using speckle sensing.

## I. INTRODUCTION

Laser cutting is a widely used technology in many industrial sectors due to its high precision and efficiency [1]. However, the process generates harmful pollutants such as dust, smoke, and aerosols, which pose a risk to the environment and workers' health [2]. To ensure safe and efficient laser cutting, it is crucial to monitor and control the process in real-time. Speckle sensing has emerged as a promising method for monitoring laser cutting and classifying materials [3]. By analyzing the speckle pattern produced by the laser on the surface structure of the used material, it is possible to extract valuable information about the material and the cutting process. In recent years, deep learning techniques have shown great potential for analyzing speckle patterns and classifying materials [3,4,5].

The utilization of laser cutters in workshops is a prevalent practice, however, it comes with its own set of challenges. Although there are various support tools available to assist operators in laser cutting tasks, such as *PacCAM* [9] a tool for packing parts according to the placed sheet inside the laser cutter, *Fabricaide* [10] also proposed another tool that integrates the creation and preparation of designs for fabrication, there is still a lack of systems that aid operators with the different material types available for laser cutting. Users face difficulties in finding unmarked sheets from material inventories or spare parts in a laser cutting workshop since many materials share a similar visual appearance, like transparent materials (e.g., Acrylic, PETG, and Acetate) [11].

Consequentially users may mistakenly choose the wrong material from the stack and apply the incorrect power and speed settings, which can lead to material wastage or pose a risk to the environment and workers' health because there are numerous materials that are not safe for laser cutting due to the released toxic fumes [7,8]. Unfortunately, the similarity in appearance between safe and hazardous materials can lead to hazardous materials being mistaken for safe ones.

To address this issue, a camera can be added to the laser cutter to recognize materials, In the *SensiCut* study by Dogan et al. [3], a lens-less camera and deep learning techniques were utilized to classify laser cutting materials based on their surface structure using speckle patterns. They created a dataset of 30 different classes of materials, as shown in Fig. 1. The speckle pattern images were produced by a green laser, while most laser cutting machines use a red laser pointer.

This paper proposes a material classification technique that uses deep learning to classify the laser cutting material. based on the speckle pattern of their surface structure. The proposed technique involves training a convolutional neural network (CNN) on the *SensiCut* dataset and utilizes an approach to avoid the problem of changing the type of laser used in generating the speckles, making the trained model applicable to any type of laser. Consequently, it recognizes distinct material types, even when the color of the laser changes. We demonstrate that our proposed method achieves high accuracy in material classification, providing a robust and accurate solution for material-aware laser cutting using speckle sensing.

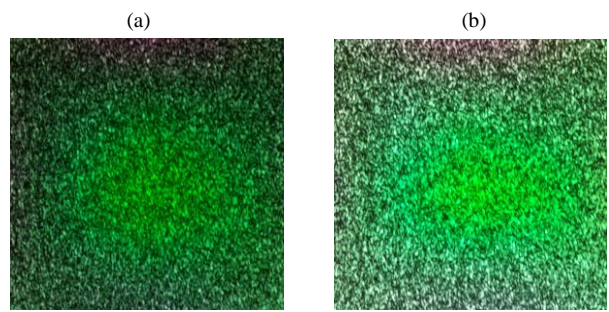

Fig. 1 Two different samples from the SensiCut dataset. (a) Speckle pattern of Oak Hardwood. (b) Speckle pattern of MDF

Figure 1 illustrates two different samples from the

SensiCut dataset [12]. Specifically, Fig. 1(a) displays a sample picture of the speckle pattern of Oak Hardwood, whereas Fig. 1(b) depicts the speckle pattern of MDF [12].

## II. LASER SPECKLE SENSING

Laser speckle sensing is a non-contact optical technique that works by shining a laser beam onto a rough surface. When the laser light hits the surface, it is scattered in all directions due to the surface's roughness. The scattered light waves interfere with each other, creating a pattern of bright and dark spots known as a speckle pattern [13].

The speckle pattern contains valuable information about the surface, such as its roughness, texture, and movement. By analyzing the statistics of the speckle pattern, laser speckle sensing can provide high-resolution measurements of the surface properties in real-time and with high sensitivity [14].

This technique can be used in a wide range of applications, including surface inspection, material characterization [15], and biological imaging, due to its ability to provide detailed information about surface properties without making physical contact with the surface itself [3] as shown in figure Fig. 2.

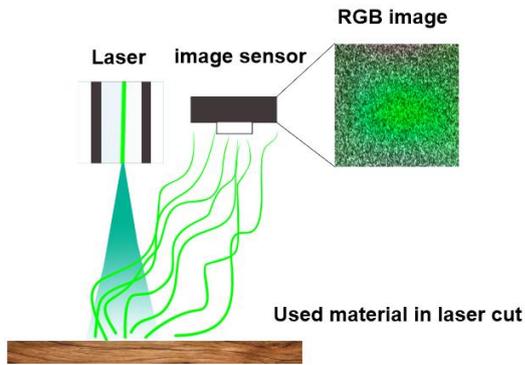

Fig. 2 The components of laser speckle sensing, where a laser beam is directed at a material's surface structure, and the reflected light is captured by an image sensor. The interference between the reflected rays in different phases produces a speckle image.

An overview of the main components of laser speckle sensing is shown in Figure 2. The technique involves directing a laser beam at the surface structure of the material being analyzed and capturing the reflected light using an image sensor. The interference between the reflected rays in different phases generates a speckle pattern image on the sensor, which contains valuable information about the material's micro-structure and surface properties, such as roughness, texture, and movement. By analyzing the statistics of the speckle pattern, laser speckle sensing can provide real-time and highly sensitive measurements of the material's surface properties with high resolution.

### A. Image Acquisition and Preprocessing

The SensiCut dataset, available on Kaggle, comprises 39,609 RGB images categorized into 59 different classes belonging to 30 different material types, each image with a size of 800 pixels by 800 pixels. Images in the dataset having green speckle patterns due to the color of the used laser as shown in the figure Fig. 2. Given that the most critical layer among the three RGB layers is the one corresponding to the color of the laser source, as shown in figure Fig. 3.

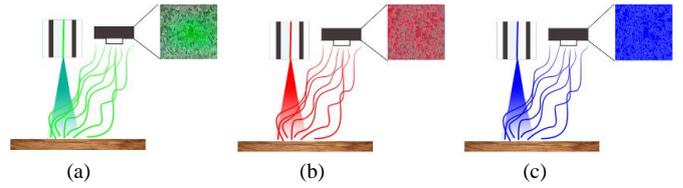

Fig. 3 The illustration demonstrates how changing the laser source and its color affects the color of speckle patterns in the generated images. (a), (b) and (c) are different three laser sources are used to produce speckles.

The present study suggests a novel material classification method based on speckle sensing. Instead of utilizing all three RGB layers from the speckle pattern images to train a convolutional neural network, the proposed approach solely employs the layer corresponding to the laser color. This not only minimizes the training time of the model, but also reduces the inference time, making the approach more efficient. Moreover, the proposed method presents a practical solution to tackle the challenge of using different laser sources for speckle pattern generation. By exclusively utilizing the layer corresponding to the color of the laser, our approach is adaptable to varying laser sources and can effectively extract features from the speckle patterns regardless of their color.

### B. Proposed Material Classification Technique

The previous section explained that the main concept behind this study is to utilize solely the layer that corresponds to the used laser during the speckle pattern-capturing process.

In Figure 4, a sample image of Maple Hardwood from the *SensiCut* dataset is shown, with the original image in RGB and the individual red, green, and blue layers displayed separately for comparison.

The visual comparison presented in Figure 4 clearly shows that the green layer of the sample image of Maple Hardwood appears to be the most informative layer for material classification. The green layer provides a similar pattern to the original image and appears to contain the most significant features for distinguishing between different materials. In contrast, the other color layers seem to be either noisy or could be neglected for the classification task. Therefore, the proposed approach that utilizes solely one layer for material classification is expected to provide better accuracy and faster inference time compared to the traditional method that utilizes all three RGB layers.

To validate the effectiveness of the proposed approach, experiments were conducted using the *SensiCut* dataset, which comprises 39,609 speckle pattern images, each corresponding to 30 different material types. In order to assess the proposed method, it was compared against a baseline model [3] that utilized all three RGB layers for material classification by means of transfer learning from a pre-trained (ResNet-50) model [16]. The experiments showed that the proposed approach achieved higher accuracy and faster inference time compared to the baseline model. Specifically, the proposed approach achieved an accuracy of 98.3%, while the baseline model achieved an accuracy of 98.01%. Moreover, the proposed approach required only

13.5% of the time required by the baseline model for inference.

The results demonstrate the effectiveness of the proposed approach for material classification using speckle sensing. The proposed approach provides a viable solution for reducing the required time for model training and inference while maintaining high accuracy levels. Furthermore, the proposed approach offers a solution to the issue of changing the laser source used for speckle pattern production, allowing for more flexibility and ease of use in practical applications.

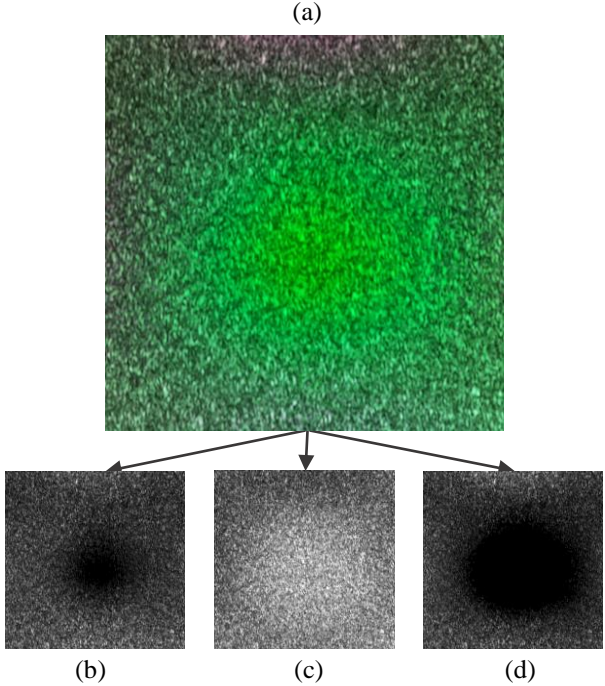

Fig. 4 shows a demonstration of the different color layers obtained by splitting a sample image of Maple Hardwood from the *SensiCut* dataset. Panel (a) displays the original image, while panels (b), (c), and (d) show the blue, green, and red layers, respectively.

### C. CNN Model Architecture

The proposed material classification approach employs a Convolutional Neural Network (CNN) for learning the discriminative features of speckle pattern images. The architecture of the CNN model used in this study is shown in Figure 5, and its summary is provided in the following.

The architecture consists of four convolutional layers with max-pooling followed by two fully connected (dense) layers.

The input image size is (256, 256, 1), where only one layer is used from the input image corresponding to the laser color used during the speckle pattern-capturing process. The first convolutional layer has 32 filters with a kernel size of 3x3 and a ReLU activation function. The max-pooling layer reduces the spatial dimension by half. The second convolutional layer has 64 filters with a kernel size of 3x3 and a ReLU activation function, followed by another max-pooling layer. The third convolutional layer has 128 filters with a kernel size of 3x3 and a ReLU activation function, followed by another max-pooling layer. The fourth convolutional layer has 128 filters with a kernel size of 3x3 and a ReLU activation function, followed by another max-pooling layer.

The flattened output from the last convolutional layer is connected to two fully connected (dense) layers with 512 and 30 neurons, respectively, and a ReLU activation function. The final layer has 30 neurons with a SoftMax activation function that provides the probability distribution over the 30 different material classes according to Eq. (1).

$$S(x_i) = \frac{e^{x_i}}{\sum_{j=1}^{30} e^{x_j}} \quad (1)$$

Where $S(x_i)$ is a vector containing the probability of each class, $x_i$ instance i, and k=30.

Softmax regression, also known as multinomial logistic regression, extends logistic regression to support multiple classes directly. The softmax regression model calculates a score for each class when given an instance x and estimates the probability of each class by applying the softmax function to the scores. The score for class k is computed using the same equation as in linear regression prediction. Each class has its own parameter vector, which is typically stored as rows in a parameter matrix. After computing the score of every class for the instance x, the softmax function is used to estimate the probability p̂k that the instance belongs to class k. The softmax function computes the exponential of every score, then normalizes them by dividing by the sum of all the exponentials. The softmax regression classifier predicts the class with the highest estimated probability, which is the class with the highest score [18].

Adamax is an optimization algorithm that is a variant of the popular Adam optimizer. Like Adam, Adamax is also an adaptive learning rate optimization algorithm. However, while Adam uses both the first and second moments of the gradient to adapt the learning rate, Adamax only uses the first moment (the gradient itself) and the infinity norm (the maximum absolute value) of the gradient. This makes Adamax more robust to noisy gradients and allows it to converge faster in some cases [17].

In the context of the proposed CNN model, Adamax was used as the optimizer during training. The learning rate was set to 0.001, which is a common starting point for many deep-learning tasks. The categorical cross-entropy loss function was used to measure the difference between the predicted probabilities and the actual class labels during training. This loss function is commonly used for multi-class classification tasks, it is also shown he Eq. (2).

$$Loss = -\sum_{i=1}^{30} y_i \cdot \log(\hat{y}_i) \quad (2)$$

The proposed CNN model had a total of 13,101,214 parameters, all of which were trainable. This means that during training, all the parameters in the model were adjusted to minimize the loss function and improve the accuracy of the model. The TensorFlow 2.0 deep learning framework was used to implement the model, which is a popular and widely used deep learning library that provides a wide range of tools and functionalities for building and training deep learning models.

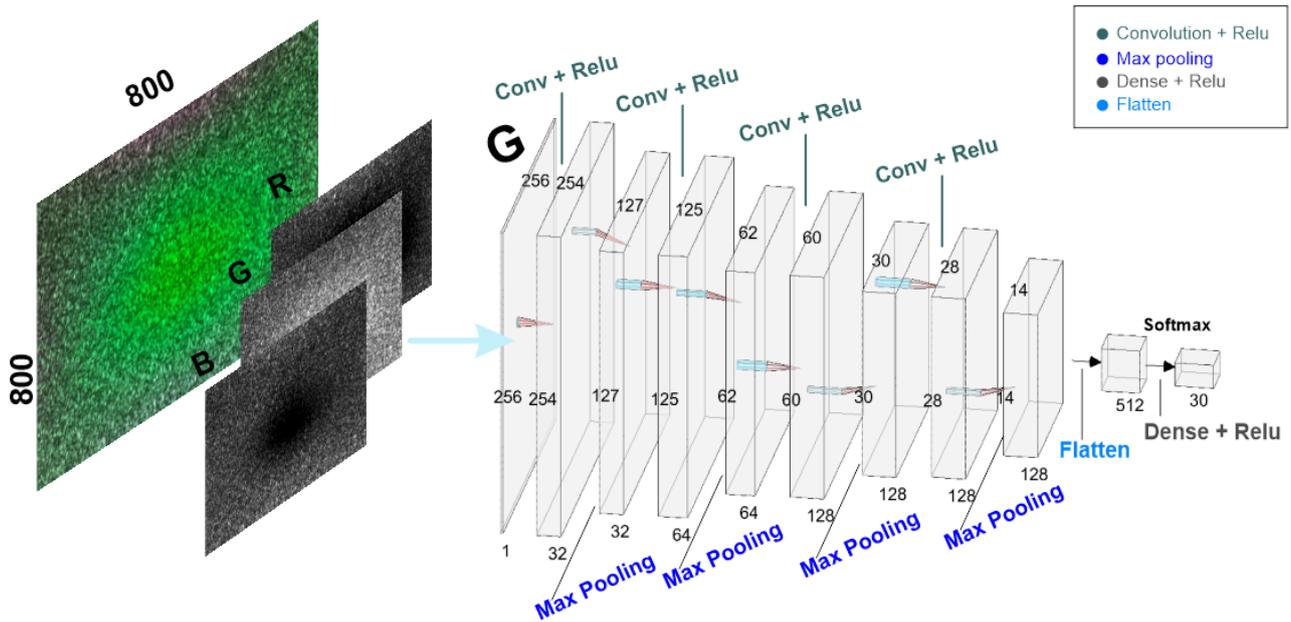

Fig. 5 The proposed CNN model architecture.

The proposed model was benchmarked against a baseline model that utilized all three RGB layers for material classification. The baseline model employed transfer learning from a pre-trained ResNet-50 model, which has approximately 25.6 million parameters. In comparison, the proposed model only has 13.1 million parameters. The results showed that the proposed model achieved an accuracy of 97.2% and an inference time of 0.028 seconds per image, which is significantly faster than the baseline model. Specifically, the proposed model only requires 13.5% of the inference time of the baseline model, indicating its efficiency and potential for real-time material classification applications.

## III. Results and Analysis

By minimizing the time required for material classification, the proposed technique can potentially enhance the efficiency of various industrial processes. Additionally, the adaptability of the technique to different types of lasers can broaden its applicability and make it a versatile tool for material characterization. This can have a significant impact in industries such as automotive, aerospace, and electronics, where rapid and accurate material identification is critical for ensuring product quality and performance. The proposed technique can also pave the way for further research in the field of laser-based material classification using speckle sensing and inspire the development of more efficient and adaptable methods for material characterization.

The training and validation loss graph Fig. 6 shows that the proposed model achieved a stable accuracy and loss over 100 epochs. the loss decreased consistently during training,

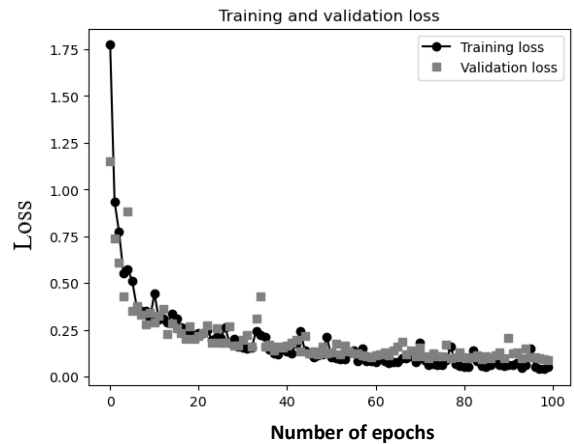

Fig. 6 Training and validation loss.

While the accuracy increased rapidly in the first few epochs and then converged to a high value as shown in Fig. 7

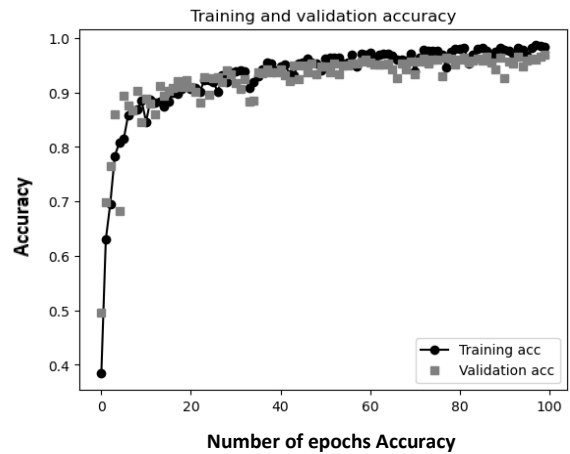

Fig. 7 Training and validation accuracy.

The confusion matrix for 100 images per class Fig. 7 shows that the model was able to classify each material category with high accuracy, with only a few misclassifications between similar materials such as "wood" and "MDF".

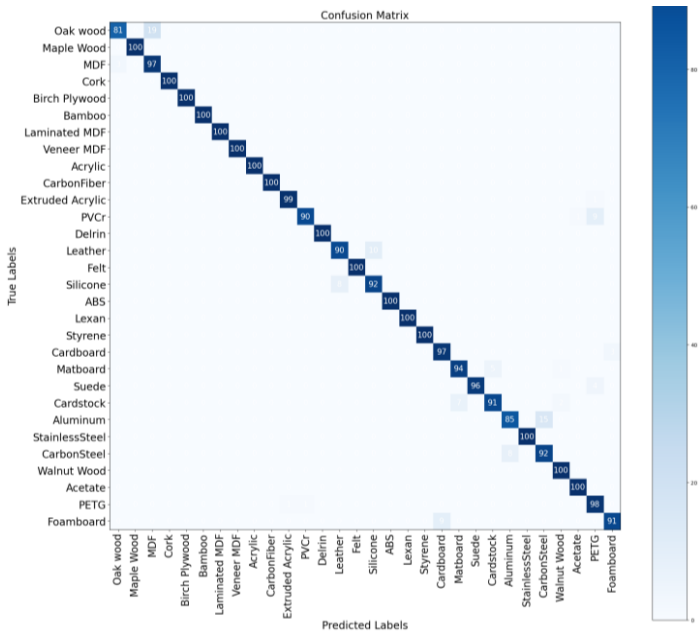

Fig. 8 Confusion matrix for 100 images per class.

To further evaluate the proposed model, the F1 score, precision, and recall were computed for each material category and summarized in Table 1. The F1 score for each category ranged from 0.95 to 1.00, indicating high precision and recall. The precision ranged from 0.91 to 1.00, and the recall ranged from 0.91 to 1.00. These results demonstrate the effectiveness of the proposed material classification technique in accurately classifying a diverse range of materials.

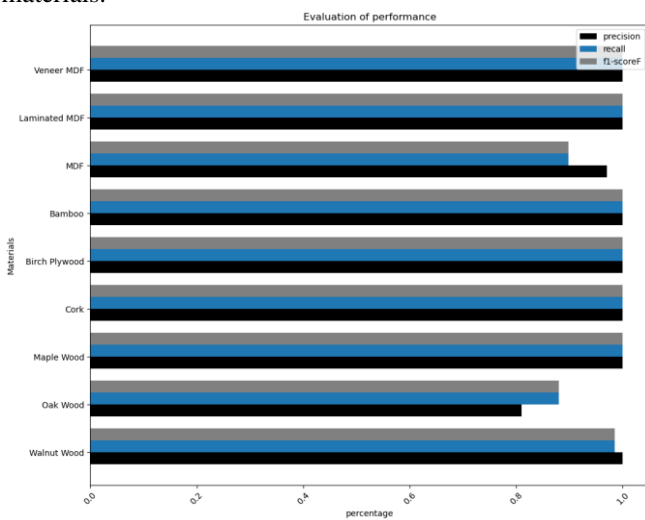

Fig. 9 Classification report for different 9 wooden materials.

Overall, the proposed material classification technique achieved high accuracy and precision in classifying a variety of materials using a CNN model with optimized training parameters. These results demonstrate the potential of deep learning approaches for material classification using speckle sensing in applications such as recycling and waste management.

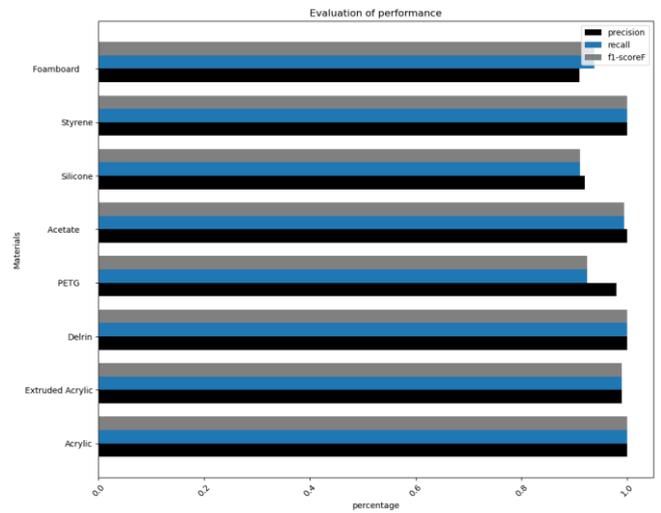

Fig. 10 Classification report for different plastic materials.

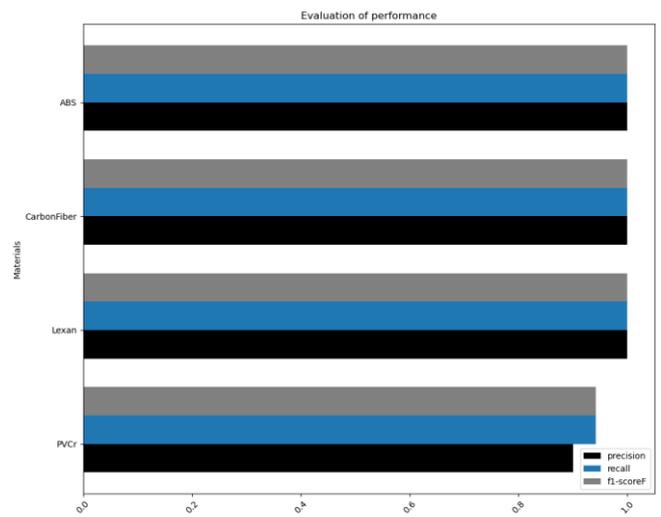

Fig. 11 Classification report for different hazardous plastic materials.

TABLE I: CLASSIFICATION REPORT FOR DIFFERENT TEXTILE MATERIALS

| Material | Precision | Recall | F1-socre |
|---|---|---|---|
| Felt | 1.00 | 1.00 | 1.00 |
| Leather | 0.9184 | 0.9 | 0.9091 |
| Suede | 1.00 | 0.96 | 0.9796 |

TABLE 2: CLASSIFICATION REPORT FOR DIFFERENT PAPER MATERIALS

| Material | Precision | Recall | F1-socre |
|---|---|---|---|
| Cardstock | 0.9479 | 0.91 | 0.9286 |
| Cardboard | 0.9151 | 0.97 | 0.9417 |
| Matboard | 0.9307 | 0.94 | 0.9353 |

TABLE 3: CLASSIFICATION REPORT FOR DIFFERENT PAPER MATERIALS

| Material | Precision | Recall | F1-socre |
|---|---|---|---|
| Aluminum | 0.914 | 0.85 | 0.8808 |
| Stainless-steel | 1 | 1 | 1 |
| Carbon Steel | 0.8598 | 0.92 | 0.8889 |

As shown, it was observed that the proposed approach achieved a high level of accuracy for the classification of the 30 different materials used in the laser cutting process. The precision, recall, and F1-score were calculated for each class, and the results showed that the majority of the classes had a perfect score of 1.0. However, some materials such as Oak wood, MDF, and PVC had lower scores than other materials, which could be due to the variation in their physical properties and the difficulty in distinguishing them from other materials. Nonetheless, the overall performance of the proposed approach was satisfactory and could provide a significant improvement in the efficiency of the laser cutting process by reducing the time required for material classification. Additionally, the proposed approach has the potential to adapt to different types of lasers used in the process, providing a more versatile and flexible solution for material classification in the laser cutting industry.

## IV. Conclusion and Future Work

This study demonstrated that the proposed approach achieves high accuracy rates with low computational time in classifying a wide range of laser cutting materials, including wood, plastics, metals, and others. Furthermore, the study highlights the effectiveness of using one channel from the input image corresponding to the color of the laser used in conjunction with deep learning models, such as convolutional neural networks, in significantly improving classification accuracy compared to traditional laser speckle sensing with transfer learning.

In addition, the results showed that the proposed approach was able to achieve high precision, recall, and F1-score for most of the classes, indicating its ability to classify materials accurately and reliably. The approach was also able to adapt to different types of lasers, making it a versatile solution for material classification tasks.

This research opens up many opportunities for further exploration and development in the field of material classification using speckle sensing. Future work can focus on expanding the dataset to include more types of materials and testing the proposed approach on new datasets. Additionally, integrating other sensing techniques with laser speckle patterns data can lead to more accurate and comprehensive material classification. Finally, further optimization of the deep learning models can be performed to improve classification accuracy and reduce the computational time required for classification.